# From imprecise probability assessments to conditional probabilities with quasi additive classes of conditioning events

**Giuseppe Sanfilippo**
Dipartimento di Scienze Statistiche e Matematiche "S. Vianelli",
University of Palermo, Italy
giuseppe.sanfilippo@unipa.it

## Abstract

In this paper, starting from a *generalized coherent* (i.e. avoiding uniform loss) interval-valued probability assessment on a finite family of conditional events, we construct conditional probabilities with *quasi additive* classes of conditioning events which are consistent with the given initial assessment. Quasi additivity assures coherence for the obtained conditional probabilities. In order to reach our goal we define a finite sequence of conditional probabilities by exploiting some theoretical results on g-coherence. In particular, we use solutions of a finite sequence of linear systems.

## 1 Introduction

In many applications of Artificial Intelligence we need to reason with uncertain information under vague or partial knowledge. A possible approach to uncertain reasoning can be based on imprecise probabilistic assessments on a family of conditional events which has no particular algebraic structure. In such a case a general framework is obtained by using suitable generalizations of the coherence principle of de Finetti (de Finetti [1974]), or similar principles adopted for lower and upper probabilities. A further advantage of using these approaches is given by the possibility of looking at the conditional probability $P(E|H)$ as a primitive concept, with no need of assuming that the probability of the conditioning event $H$ be positive. In the past, the coherence-based approach to probabilistic reasoning has covered several natural semantics for conditionals, included the ones used in default reasoning such as System P (see, e.g., Gilio [2002], Biazzo et al. [2005], Gilio and Sanfilippo [2011]). In this paper we adopt a notion of generalized coherence called *g-coherence* given in Biazzo and Gilio [2000] (see also Coletti [1994], Gilio [1995b], Biazzo et al. [2005]). The notion of g-coherence is weaker than the notion of coherence given for lower and upper probabilities (see Walley [1991], Williams [2007]) and is equivalent (Biazzo and Gilio [2002]) to the property of "avoiding uniform loss" given in Walley [1991, 1997]. Using some algorithms, a g-coherent probability assessment can be corrected obtaining a coherent assessment. We recall that, given a coherent (precise) assessment $\mathcal{P}$ on an arbitrary family $\mathcal{F}$ of conditional events, there always exists a coherent extension of $\mathcal{P}$ which is a conditional probability (Regazzini [1985]), see also Holzer [1985], Rigo [1988]). Conversely, given a conditional probability $P$ on $\mathcal{E} \times \mathcal{X}$, where $\mathcal{E}$ is an algebra and $\mathcal{X}$ is a nonempty subset of $\mathcal{E} \setminus \{\emptyset\}$, in Rigo [1988] is shown that a suitable condition given in Császár [1955] (called Csàszàr's condition) is necessary and sufficient for coherence of $P$. If no restrictions are supposed regarding the class of conditioning events $\mathcal{X}$ the function $P$ could be not coherent (Gilio and Spezzaferri [1992], Gilio [1995a], Coletti and Scozzafava [2002]). Moreover, if $\mathcal{X}$ has a particular structure, for instance $\mathcal{X}$ is additive (Holzer [1985], Coletti and Scozzafava [2002]) or quasi-additive (Gilio [1989]), then $P$ is coherent. As quasi-additivity property is weaker than additivity it is of some interest to check if a given coherent assessment $\mathcal{P}$ could be extended as a conditional probability with the class of conditioning events quasi-additive. We remark that the such conditional probabilities may be useful because in many applications in order to take a decision we need to choose a precise and (possibly complete) probability assessment. Hence, the kind of problems studied in this paper can help the decision maker to 'integrate knowledge' by computing a probability distribution consistent with the 'incomplete or vague' initial information quantified by the imprecise assessment.

In what follows, starting from a g-coherent interval-valued probability assessment $\mathcal{A}$ on a finite family $\mathcal{F}$ of conditional events we first determine the associated interval-valued assessment $\mathcal{A}^*$ on $\mathcal{F}$ coherent in the sense of Walley. Then, after recalling the notion of finitely additive conditional probability and for the set of conditioning events the property of quasi additivity, we construct a sequence of conditional probabilities with quasi additive classes of conditioning events. We extend the conditional probabilities of this sequence using the same (quasi additive) classes of conditioning events. Thus, we are able to define a conditional probability $P_{01\ldots k}$ on $\mathcal{E} \times \mathcal{X}$, where $\mathcal{E}$ is the algebra generated by $\mathcal{F}$ and $\mathcal{X}$ is a quasi additive class which coincides with the union of the previous classes of conditioning events. Moreover, $\mathcal{X}$ contains the initial set of conditioning events of $\mathcal{F}$. Finally, we observe that $P_{01\ldots k}$ is a coherent conditional probability on $\mathcal{F}$ consistent with $\mathcal{A}$. By this procedure one can obtain a large (potentially infinite) number of coherent conditional probabilities. The paper is organized as follows. In Section 2 we first give some preliminary notions and results on generalized coherence. Then, we recall the concepts of conditional probability and quasi additive class of events. Finally we recall a sufficient condition for coherence of conditional probability. In Section 3 we show how a g-coherence assessment, using some algorithms, can be corrected obtaining a coherent assessment. In Section 4 given a g-coherent interval-valued assessment $\mathcal{A}$ on $\mathcal{F}$ we construct a finite sequence $P_0, P_1, \ldots, P_k$ of conditional probabilities with quasi additive families of conditioning events. In Section 5 we introduce a new sequence of conditional probabilities $P_0^0, P_1^0, \ldots, P_k^0$ with quasi additive families of conditioning events such that, for each $i$, the probability $P_i^0$ is a particular extension of $P_i$. We also show that $P_i^0$ is coherent. In Section 6 we define a conditional probability $P_{01\ldots k}$ with a quasi additive class of conditioning events which contains all conditioning events on $\mathcal{F}$. We show that $P_{01\ldots k}$ is coherent and its restriction on $\mathcal{F}$ is a coherent assessment which is consistent with $\mathcal{A}$. In Section 7 we give a characterization theorem of coherent precise assessments and a relevant way of introducing quasi additive classes. Finally, we give some conclusions.

# 2 Preliminary notions and results

In this section we set up notation and terminology. We also recall some basic concepts and results, related with the checking of g-coherence and propagation of conditional probability bounds. Next, we recall the definition of conditional probabilities. Finally, we recall the notion of quasi-additive class of events and a sufficient condition for coherence which will be used in the next sections.

## 2.1 Notations

For each integer $n$ we set $J_n = \{1, 2, \ldots, n\}$. Moreover, we denote respectively by $\Omega$ the sure event, by $\emptyset$ the impossible event and the empty set, by $E^c$ the negation of the event $E$, by $E \vee H$ the disjunction of $E$ and $H$ and by $E \wedge H$ (or simply by $EH$) the conjunction of $E$ and $H$. The logical implication from $E$ to $H$, namely *if $E$ is true, then also $H$ is true*, is denoted by $E \subseteq H$.

## 2.2 Constituents

Given any pair of events $E$ and $H$, with $H \neq \emptyset$, we look at the conditional event $E|H$ as a three-valued logical entity which is true, or false, or void, according to whether $EH$ is true, or $E^cH$ is true, or $H^c$ is true. In the setting of coherence, agreeing to the betting metaphor, if you assess $P(E|H) = p$, then you agree to pay an amount $p$, by receiving 1, or 0, or $p$, according to whether $E|H$ is true, or $E|H$ is false, or $E|H$ is void (bet called off). Given a family of conditional events $\mathcal{F}_{J_n} = \{E_i|H_i, i \in J_n\}$, we set $\mathcal{H}_{J_n} = \bigvee_{j \in J_n} H_j$. For each $r \in \{0, 1, \ldots 3^n - 1\}$, we denote by $(r_n r_{n-1} \cdots r_2 r_1)$ its ternary representation, such as $r = r_1 3^0 + r_2 3^1 + \cdots + r_n 3^{r_n - 1}$, and by $R_r$ the event defined as $R_r = R_1^{r_1} R_2^{r_2} \cdots R_n^{r_n}$, where

$$R_i^{r_i} = \begin{cases} E_i H_i, & \text{if } r_i = 1, \\ E_i^c H_i, & \text{if } r_i = 0, \\ H_i^c, & \text{if } r_i = 2. \end{cases}$$

Then, we introduce the set

$$\mathcal{C}_{J_n} = \{R_r \neq \emptyset, r \in \{0, 1, 2, \ldots, 3^n - 1\}\}$$

containing every $R_r$ that is not impossible. As some conjunction $R_r$ may be impossible, the cardinality of $\mathcal{C}_{J_n}$ is less than or equal to $3^n$. We call *constituents* or *possible worlds* associated with $\mathcal{F}_{J_n}$ the elements of $\mathcal{C}_{J_n}$. Obviously $\mathcal{C}_{J_n}$ is a partition of $\Omega$, that is
1. $\bigvee_{C \in \mathcal{C}_{J_n}} C = \Omega$,
2. $C' \wedge C'' = \emptyset$ for each $C', C'' \in \mathcal{C}_{J_n}$ with $C' \neq C''$.
With each event $E$ we associate the following subset of $\mathcal{C}_{J_n}$ $C_{J_n}(E) = \{C \in \mathcal{C}_{J_n} :\ C \subseteq E\}$. Notice that for each pair of events $E_i H_i$,$H_i$, since $E_i H_i \subseteq H_i \subseteq \mathcal{H}_{J_n}$, it follows that $C_{J_n}(E_i H_i) \subseteq C_{J_n}(H_i) \subseteq C_{J_n}(\mathcal{H}_{J_n})$.

**Example 1.** Let $\mathcal{F}_{J_3} = \{E_1|H_1, E_2|H_2, E_3|H_3\} = \{ABC|D, B|AC, C|AB\}$ be a family of three conditional events. The set $\mathcal{C}_{J_3}$ of the constituents associated with $\mathcal{F}_{J_3}$ is

$$\mathcal{C}_{J_3} = \{R_6, R_8, R_{13}, R_{14}, R_{18}, R_{20}, R_{24}, R_{26}\} = \\ = \{C_1, C_2, C_3, C_4, C_5, C_6, C_7, C_8\},$$

where

$$C_1 = ABC^cD,\ C_2 = ABC^cD^c,\ C_3 = ABCD, \\ C_4 = ABCD^c,\ C_5 = AB^cCD,\ C_6 = AB^cCD^c, \\ C_7 = (A^c \vee B^cC^c)D,\ C_8 = (A^c \vee B^cC^c)D^c.$$

Moreover, we have that $\mathcal{H}_{J_3} = \{D \vee AC \vee AB\}$ and

$$\mathcal{C}_{J_3}(\mathcal{H}_{J_3}) = \mathcal{C}_{J_3} \setminus \{C_8\} = \{C_1, C_2, \ldots, C_7\}. \qquad (1)$$

### 2.3 Coherence and g-coherence

Given an arbitrary family of conditional events $\mathcal{F}$ and a real function $\mathcal{P}$ on $\mathcal{F}$, for every $n \in \mathbb{N}$, let $\mathcal{F}_{J_n} = \{E_i|H_i, i \in J_n\}$ be a subfamily of $\mathcal{F}$ and $\mathcal{P}_{J_n}$ the vector $(p_i, i \in J_n)$, where $p_i = \mathcal{P}(E_i|H_i)$. We use the same symbols for events and their indicator. Then, considering the random gain

$$G_{J_n} = \sum_{i \in J_n} s_i H_i (E_i - p_i),$$

with $s_i, i \in J_n$, arbitrary real numbers, we denote by $G_{J_n}|\mathcal{H}_{J_n}$ the restriction of $G_{J_n}$ to $\mathcal{H}_{J_n}$. Then, based on the betting scheme, we have

**Definition 1.** The function $\mathcal{P}$ is said coherent iff

$$\max G_{J_n}|\mathcal{H}_{J_n} \geq 0, \, \forall n \geq 1, \forall \mathcal{F}_{J_n} \subseteq \mathcal{F}, \forall s_i \in \mathbb{R}, i \in J_n.$$

Given an interval-valued probability assessment $\mathcal{A}_{J_n} = ([a_1, b_1], \ldots, [a_n, b_n])$ on a family $\mathcal{F}_{J_n} = \{E_i|H_i, i \in J_n\}$ we adopt the following condition of generalized coherence (*g-coherence*) given in Biazzo and Gilio [2000].

**Definition 2.** The interval-valued probability assessment $\mathcal{A}_{J_n}$ on $\mathcal{F}_{J_n}$ is said g-coherent if and only if there exists a precise coherent assessment $\mathcal{P}_{J_n} = (p_i, \, i \in J_n)$ on $\mathcal{F}_{J_n}$, with $p_i = P(E_i|H_i)$, which is consistent with $\mathcal{A}_{J_n}$, that is such that $a_i \leq p_i \leq b_i$ for each $i \in J_n$.

Given an interval-valued probability assessment $\mathcal{A}_{J_n}$ on $\mathcal{F}_{J_n}$, we denote by $(\mathcal{S}_{J_n})$ the following system, associated with the pair $(\mathcal{F}_{J_n}, \mathcal{A}_{J_n})$, with nonnegative unknown $\underline{\lambda} = (\lambda_C, \, C \in \mathcal{C}_{J_n}(\mathcal{H}_{J_n}))$:

$$(\mathcal{S}_{J_n}) \begin{cases} \displaystyle\sum_{C \in \mathcal{C}_{J_n}(E_i H_i)} \lambda_C \leq b_i \cdot \sum_{C \in \mathcal{C}_{J_n}(H_i)} \lambda_C \, , \forall i \in J_n \\ \displaystyle\sum_{C \in \mathcal{C}_{J_n}(E_i H_i)} \lambda_C \geq a_i \cdot \sum_{C \in \mathcal{C}_{J_n}(H_i)} \lambda_C \, , \forall i \in J_n \\ \displaystyle\sum_{C \in \mathcal{C}_{J_n}(\mathcal{H}_{J_n})} \lambda_C = 1 \, , \quad \lambda_C \geq 0, \forall C \in \mathcal{C}_{J_n}(\mathcal{H}_{J_n}). \end{cases}$$

In an analogous way, given a subset $J$ of $J_n$, we denote by $(\mathcal{F}_J, \mathcal{A}_J)$ the pair corresponding to $J$, by $(\mathcal{S}_J)$ the system associated with $(\mathcal{F}_J, \mathcal{A}_J)$ and by $\mathcal{C}_J$ the set of constituents associated with $\mathcal{F}_J$.
We recall a result given in Gilio [1995a], where the notion of g-coherence (see Biazzo and Gilio [2000]) was simply named coherence.

**Theorem 1.** *The interval-valued probability assessment $\mathcal{A}_{J_n}$ on $\mathcal{F}_{J_n}$ is g-coherent if and only if, for every $J \subseteq J_n$, the system $(\mathcal{S}_J)$ is solvable.*

### 2.4 Conditional probabilities

Given an algebra of events $\mathcal{E}$ and a non empty subfamily $\mathcal{X}$ of $\mathcal{E}$, with $\emptyset \notin \mathcal{X}$, a (finitely-additive) *conditional probability* on $\mathcal{A} \times \mathcal{X}$ is a real-valued function $P$ defined on $\mathcal{E} \times \mathcal{X}$ satisfying the following properties (Dubins [1975], see also Rényi [1955], Császár [1955], Regazzini [1985], Coletti [1994], Coletti and Scozzafava [2002]):

- $(i)$ $P(\cdot|H)$ is a finitely additive probability on $\mathcal{E}$, for each $H \in \mathcal{X}$;
- $(ii)$ $P(H|H)$=1, for each $H \in \mathcal{X}$;
- $(iii)$ $P(E_1 E_2|H) = P(E_2|E_1 H) P(E_1|H)$, for every $E_1, E_2, H$, with $E_1 \in \mathcal{E}$, $E_2 \in \mathcal{E}$, $H \in \mathcal{X}$ and $E_1 H \in \mathcal{X}$.

As in Rényi [1955] we do not suppose any restrictions regarding the class of conditioning events $\mathcal{X}$ of the conditional probability $P$. In Dubins [1975], in order to define a finitely-additive conditional probability, it is required that $\mathcal{X} \cup \{\emptyset\}$ must be a subalgebra of $\mathcal{E}$.

### 2.5 Quasi additivity and coherence

Let $P$ be a conditional probability on $\mathcal{E} \times \mathcal{X}$, the family $\mathcal{X}$ of conditioning events is said a $P-$*quasi-additive* class (or quasi additive w.r.t. $P$) if, for every $H_1 \in \mathcal{X}, H_2 \in \mathcal{X}$, there exists $K \in \mathcal{X}$ such that (Császár [1955])

$$\begin{array}{ll} (i) & H_1 \vee H_2 \subseteq K, \\ (ii) & P(H_1|K) + P(H_2|K) > 0. \end{array} \qquad (2)$$

We observe that, given any conditional probability $P$ on $\mathcal{E} \times \mathcal{X}$, if $\mathcal{X}$ is additive or $\mathcal{X} \cup \{\emptyset\}$ is a subalgebra of $\mathcal{E}$, then $\mathcal{X}$ is $P-$quasi additive. Some sufficient conditions for coherence of a conditional probability are given in Gilio [1989] from which we recall the following result

**Theorem 2.** *If $P$ is a conditional probability on $\mathcal{E} \times \mathcal{X}$, with $\mathcal{E}$ algebra and $\mathcal{X}$ quasi additive class w.r.t. $P$, then $P$ is coherent.*

## 3 From g-coherent to coherent lower/upper probabilities

Let $\mathcal{A}_{J_n} = ([a_i, b_i], \, i \in J_n)$ be a g-coherent interval-valued probability assessment on $\mathcal{F}_{J_n} = \{E_i|H_i, \, i \in J_n\}$. Of course, we can assume that, if $E_i|H_i \in \mathcal{F}_{J_n}$, then $E_i^c|H_i \notin \mathcal{F}_{J_n}$. Moreover, if we replace each upper bound $P(E_i|H_i) \leq b_i$ by the equivalent lower bound $P(E_i^c|H_i) \geq 1 - b_i$, the assessment $\mathcal{A}_{J_n}$ on $\mathcal{F}_{J_n}$ can be seen as a lower probability $(a_i, 1 - b_i, \, i \in J_n)$ on the family $\{E_i|H_i, E_i^c|H_i, \, i \in J_n\}$. Then, using a suitable alternative theorem, it can be shown that g-coherence and avoiding uniform loss (AUL) property of lower and upper probabilities (Walley [1991]) are equivalent (see Biazzo and Gilio [2002], Theorem 10).
Given a further conditional event $E_{n+1}|H_{n+1}$, as it can be verified (Biazzo and Gilio [2000]), there exists a suitable interval $[p_\circ, p^\circ]$ such that

**Theorem 3.** *The interval-valued assessment* $\mathcal{A}_{J_{n+1}} = ([a_i, b_i],\ i \in J_{n+1})$ *with* $a_{n+1} = b_{n+1} = p_{n+1}$ *on the family* $\mathcal{F}_{J_{n+1}} = \{E_i|H_i,\ i \in J_{n+1}\}$*, is g-coherent if and only if* $p_{n+1} \in [p_\circ, p^\circ]$.

Then, it immediately follows

**Theorem 4.** *Given a g-coherent interval-valued assessment* $\mathcal{A}_{J_n} = ([a_i, b_i],\ i \in J_n)$ *on the family* $\mathcal{F}_{J_n} = \{E_i|H_i,\ i \in J_n\}$*, the extension* $[a_{n+1}, b_{n+1}]$ *of* $\mathcal{A}_{J_n}$ *to a further conditional event* $E_{n+1}|H_{n+1}$ *is g-coherent if and only if* $[a_{n+1}, b_{n+1}] \cap [p_\circ, p^\circ] \neq \emptyset$ .

The values $p_\circ, p^\circ$ can be determined by exploiting a suitable algorithm given in Biazzo and Gilio [2000]. By the same algorithm, starting with a g-coherent assessment $\mathcal{A}_{J_n}$ on $\mathcal{F}_{J_n}$, we can make the "least-committal" correction (see Pelessoni and Vicig [1998]) of $\mathcal{A}_{J_n}$.
In this way, we obtain the coherent (lower and upper) probability $\mathcal{A}^*_{J_n}$ on $\mathcal{F}_{J_n}$ which would be produced by applying the natural extension principle proposed in Walley [1991].
To determine $\mathcal{A}^*_{J_n}$, we just need to apply $n$ times such algorithm, by replacing each time $E_{n+1}|H_{n+1}$ by $E_j|H_j$, $j \in J_n$, using as probabilistic constraints on the conditional events of $\mathcal{F}_{J_n}$ the g-coherent assessment $\mathcal{A}_{J_n}$.

**Example 1** (continued)
Let $\mathcal{A}_{J_3} = ([\frac{1}{2}, 1], [0, \frac{1}{2}], [\frac{1}{3}, \frac{2}{3}])$ be an imprecise probability assessment on $\mathcal{F}_{J_3} = \{ABC|D, B|AC, C|AB\}$. By applying the algorithm for checking g-coherence given in Biazzo and Gilio [2000] it can be proved that $\mathcal{A}_{J_3}$ is g-coherent. Moreover, $\mathcal{A}_{J_3}$ is coherent [1] and then $\mathcal{A}_{J_3} = \mathcal{A}^*_{J_3}$.

# 4 Construction of classes of conditional probabilities with quasi additive families of conditioning events

In this section, starting with a finite interval-valued probability assessment, we will construct classes of conditional probabilities with quasi additive families of conditioning events. Quasi additivity will assure coherence for the obtained conditional probabilities.

**Remark 1.** We observe that $\mathcal{C}_{J_n} = \mathcal{C}_{J_n}(\mathcal{H}_{J_n}) \cup \{\mathcal{H}^c_{J_n}\}$. In particular, $\mathcal{C}_{J_n} = \mathcal{C}_{J_n}(\mathcal{H}_{J_n})$, if $\mathcal{H}_{J_n} = \Omega$. Moreover, the system $(\mathcal{S}_{J_n})$ is solvable if and only if the following system is solvable

$$
(\mathcal{S}^*_{J_n})\begin{cases} \sum_{C \in \mathcal{C}_{J_n}(E_iH_i)} \lambda_C \leq b_i \cdot \sum_{C \in \mathcal{C}_{J_n}(H_i)} \lambda_C, \forall\, i \in J_n \\ \sum_{C \in \mathcal{C}_{J_n}(E_iH_i)} \lambda_C \geq a_i \cdot \sum_{C \in \mathcal{C}_{J_n}(H_i)} \lambda_C, \forall\, i \in J_n \\ \sum_{C \in \mathcal{C}_{J_n}(\mathcal{H}_{J_n})} \lambda_C > 0\,,\ \ \lambda_C \geq 0\ \ \forall\, C \in \mathcal{C}_{J_n}. \end{cases}
$$

In fact, given a solution $\underline{\lambda}^* = (\lambda^*_C, C \in \mathcal{C}_{J_n})$ of $(\mathcal{S}^*_{J_n})$, the vector $\underline{\lambda}' = (\lambda'_C, C \in \mathcal{C}_{J_n}(\mathcal{H}_{J_n}))$, defined by

$$
\lambda'_C = \frac{\lambda^*_C}{\sum_{C \subseteq \mathcal{H}_{J_n}} \lambda^*_C}\,,\ \forall\, C \in \mathcal{C}_{J_n}(\mathcal{H}_{J_n})\,,
$$

is a solution of $(\mathcal{S}_{J_n})$. Conversely, given a solution $\underline{\lambda}' = (\lambda'_C, C \in \mathcal{C}_{J_n}(\mathcal{H}_{J_n}))$ of $(\mathcal{S}_{J_n})$, any vector $\underline{\lambda}^* = (\lambda^*_C, C \in \mathcal{C}_{J_n})$, with

$$
\lambda^*_C = \alpha \lambda'_C\,,\ \ \forall C \in \mathcal{C}_{J_n}(\mathcal{H}_{J_n})\,,\ \ \alpha > 0\,,\ \ \lambda^*_{\mathcal{H}^c_{J_n}} \geq 0\,,
$$

is a solution of $(\mathcal{S}^*_{J_n})$. Of course, the variable $\lambda_{\mathcal{H}^c_{J_n}}$ in $(\mathcal{S}^*_{J_n})$ disappears when $\mathcal{H}^c_{J_n} = \emptyset$.
In what follows, we will use system $(\mathcal{S}^*_{J_n})$ in the equivalent formulation given below

$$
(\mathcal{S}^*_{J_n})\begin{cases} \sum_{C \in \mathcal{C}_{J_n}(E_iH_i)} \lambda_C \leq b_i \cdot \sum_{C \in \mathcal{C}_{J_n}(H_i)} \lambda_C, \forall i \in J_n \\ \sum_{C \in \mathcal{C}_{J_n}(E_iH_i)} \lambda_C \geq a_i \cdot \sum_{C \in \mathcal{C}_{J_n}(H_i)} \lambda_C, \forall i \in J_n \\ \sum_{C \in \mathcal{C}_{J_n}} \lambda_C = \sum_{C \in \mathcal{C}_{J_n}(\mathcal{H}_{J_n})} \lambda_C = 1\,, \\ \lambda_C \geq 0\ \ , \forall C \in \mathcal{C}_{J_n}. \end{cases}
$$

**Remark 2.** We denote by $\Pi$ the set of coherent precise assessments $\mathcal{P} = (p_{E|H},\ \ E|H \in \mathcal{F}_{J_n})$ on $\mathcal{F}_{J_n}$ which are consistent with $\mathcal{A}_{J_n}$. Let $\mathcal{A}^*_{J_n} = ([a^*_i, b^*_i], i \in J_n)$ be the coherent assessment associated with $\mathcal{A}_{J_n}$, computed by the procedure cited in the previous section. We recall that coherence of $\mathcal{A}^*_{J_n}$ amounts to the existence, for any given $j \in J_n$ and any $x_j \in [a^*_j, b^*_j]$, of a coherent precise probability assessment $(p_{E_i|H_i},\ \ i \in J_n)$ on $\mathcal{F}_{J_n}$, which is consistent with $\mathcal{A}^*_{J_n}$ and is such that $p_{E_j|H_j} = x_j$. We observe that, denoting by $\Pi^*$ the set of coherent precise assessments $\mathcal{P} = (p_{E|H},\ \ E|H \in \mathcal{F}_{J_n})$ on $\mathcal{F}_{J_n}$ which are consistent with $\mathcal{A}^*_{J_n}$, it holds that $\Pi = \Pi^*$.

To construct the classes of conditional probabilities associated with a given pair $(\mathcal{F}_{J_n}, \mathcal{A}_{J_n})$, we will use a suitable finite sequence $(\mathcal{F}_0, \mathcal{A}_0)$, $(\mathcal{F}_1, \mathcal{A}_1), \ldots, (\mathcal{F}_k, \mathcal{A}_k)$, with $(\mathcal{F}_0, \mathcal{A}_0) = (\mathcal{F}_{J_n}, \mathcal{A}_{J_n})$ and $\mathcal{F}_0 \supset \mathcal{F}_1 \supset \cdots \supset \mathcal{F}_k$, where $\mathcal{A}_j$ is the sub-assessment associated with $\mathcal{F}_j$.
Rather than discuss this in full generality, let us look at

[1] Coherence can also be checked by the CkC-package (Baioletti et al.) available at
`http://www.dmi.unipg.it/~upkd/paid/software.html`

$(\mathcal{F}_0, \mathcal{A}_0)$. Let $\mathcal{C}_0$ be the set of constituents associated with $\mathcal{F}_0$ and $\Pi_0$ the set of coherent precise assessments $\mathcal{P}$ on $\mathcal{F}_0$ which are consistent with $\mathcal{A}_0$ . Then, given a precise coherent assessment $\mathcal{P}_0 = (p^{(0)}_{E|H},\ \ E|H \in \mathcal{F}_0) \in \Pi_0$ on $\mathcal{F}_0$ we consider the following system $(\mathcal{S}_0)$ in the unknowns $\underline{\lambda} = (\lambda_C, C \in \mathcal{C}_0)$ associated with $(\mathcal{P}_0, \mathcal{F}_0)$

$$(\mathcal{S}_0)\begin{cases} \sum\limits_{C\in\mathcal{C}_0(EH)} \lambda_C = p_{E|H}\cdot \sum\limits_{C\in\mathcal{C}_0(H)} \lambda_C,\ \forall E|H \in \mathcal{F}_0 \\ \sum\limits_{C\in\mathcal{C}_0} \lambda_C = \sum\limits_{C\in\mathcal{C}_0(\mathcal{H}_0)} \lambda_C = 1\,,\ \ \lambda_C \geq 0\ \ \forall C \in \mathcal{C}_0, \end{cases}$$

where $\mathcal{H}_0 = \bigvee_{E|H\in\mathcal{F}_0} H$.

We observe that $\mathcal{P}_0$ is a particular interval-valued probability assessment on $\mathcal{F}_0$ where each upper bound coincides with the respective lower bound. Therefore, based on Theorem 1 and Remark 1, the system $(\mathcal{S}_0)$ is solvable. Denoting by $\mathcal{E}_0$ the algebra generated by the elements of $\mathcal{C}_0$, we introduce the following real function of the pair $(E, \underline{\lambda})$, with $E$ in $\mathcal{E}_0$ and $\underline{\lambda} = (\lambda_C, C \in \mathcal{C}_0)$ a vector of non-negative numbers,

$$\phi_0(E, \underline{\lambda}) = \begin{cases} \sum_{C\in\mathcal{C}_0(E)} \lambda_C, & \text{if } C_0(E) \neq \emptyset \\ 0, & \text{otherwise.} \end{cases} \tag{3}$$

Next, let $\underline{\lambda}^\star = (\lambda^\star_C : C \in \mathcal{C}_0)$ be a solution of the system $(\mathcal{S}_0)$ and $D_0 = \{H : E|H \in \mathcal{F}_0\}$ be the set of conditioning events of $\mathcal{F}_0$, we consider the following partition of $D_0$

$$D^z_0 = \{H \in D_0\ : \phi_0(H, \underline{\lambda}^\star) = 0\}, \tag{4}$$

$$D^+_0 = D_0 \setminus D^z_0 = \{H \in D_0\ : \phi_0(H, \underline{\lambda}^\star) > 0\ \}. \tag{5}$$

**Remark 3.** Note that the subset $D^+_0$ cannot be empty (that is $D^z_0 \subset D_0$). In fact, as $\underline{\lambda}^\star$ is a solution of $(S_0)$ we have

$$\sum_{H\in D_0} \phi_0(H, \underline{\lambda}^\star) \geq \sum_{C\in\mathcal{C}_0(\mathcal{H}_0)} \lambda^\star_C = \sum_{C\in\mathcal{C}_0} \lambda^\star_C = 1.$$

Hence, there exists an event $H \in D_0$ such that $\phi(H, \underline{\lambda}^\star) > 0$, that is $D^+_0 \neq \emptyset$.

Let $\mathcal{X}_0 = D^+_0 \cup \{\mathcal{H}_0\}$ be a family of conditioning events and $\underline{\lambda}^\star$ be a solution of $(S_0)$, we set

$$P_0(E|H) = \frac{\phi_0(EH, \underline{\lambda}^\star)}{\phi_0(H, \underline{\lambda}^\star)}, \forall E|H,\ \text{with } E \in \mathcal{E}_0,\ H \in \mathcal{X}_0. \tag{6}$$

Then, we have

**Theorem 5.** *The real function $P_0$ is a conditional probability on $\mathcal{E}_0 \times \mathcal{X}_0$, with $\mathcal{X}_0$ quasi-additive w.r.t. $P_0$.*

*Proof.* In order to prove that $P_0$ is a conditional probability on $\mathcal{E}_0 \times \mathcal{X}_0$ we need to show that $P_0$ satisfies properties $(i), (ii), (iii)$. Then, such a proof will be divided into three steps.

1) Let $H$ be a conditioning event in $\mathcal{X}_0$. We obviously have
$$P_0(\Omega|H) = \frac{\phi_0(H, \underline{\lambda}^\star)}{\phi_0(H, \underline{\lambda}^\star)} = 1.$$
Moreover, for every $E \in \mathcal{E}_0$ one has $P_0(E|H) \geq 0$. Finally, for every couple of incompatible events $E_1, E_2$ in $\mathcal{E}_0$, we have
$$P_0(E_1 \vee E_2|H) = \tfrac{\phi_0(E_1H\vee E_2H, \underline{\lambda}^\star)}{\phi_0(H,\underline{\lambda}^\star)} = \tfrac{\phi_0(E_1H,\underline{\lambda}^\star)}{\phi_0(H,\underline{\lambda}^\star)} + \tfrac{\phi_0(E_2H,\underline{\lambda}^\star)}{\phi_0(H,\underline{\lambda}^\star)} = P_0(E_1|H) + P_0(E_2|H).$$
Therefore, for each $H \in \mathcal{X}_0$, $P_0(\cdot|H)$ is a finitely additive probability on $\mathcal{E}_0$; that is $P_0$ satisfies property $(i)$.

2) For each conditioning event $H \in \mathcal{X}_0$ we trivially have $P_0(H|H) = 1$. Then, $P_0$ satisfies property $(ii)$.

3) Let $E_1, E_2, H$ be three events in $\mathcal{E}_0$ such that $H$ and $E_1H$ are in $\mathcal{X}_0$. Then, one has $\phi_0(H, \underline{\lambda}^\star) > 0$ and $\phi_0(E_1H, \underline{\lambda}^\star) > 0$. Therefore, we obtain
$$\begin{aligned} P_0(E_1E_2|H) &= \tfrac{\phi_0(E_1E_2H,\underline{\lambda}^\star)}{\phi_0(H,\underline{\lambda}^\star)}, \\ P_0(E_2|E_1H) &= \tfrac{\phi_0(E_1E_2H,\underline{\lambda}^\star)}{\phi_0(E_1H,\underline{\lambda}^\star)}, \\ P_0(E_1|H) &= \tfrac{\phi_0(E_1H,\underline{\lambda}^\star)}{\phi_0(H,\underline{\lambda}^\star)}. \end{aligned}$$
Hence, it is easily seen that $P_0(E_1E_2|H) = P_0(E_2|E_1H)P_0(E_1|H)$, so that $P_0$ satisfies property $(iii)$.

Therefore, $P_0$ is a conditional probability on $\mathcal{E}_0 \times \mathcal{X}_0$. Next, we prove that $\mathcal{X}_0$ is a $P_0-$quasi additive class. Given two conditioning events $H_1, H_2 \in \mathcal{X}_0$ we have $H_1 \vee H_2 \subseteq \mathcal{H}_0$. In addition, as $\phi_0(H_1, \underline{\lambda}^\star) > 0$ and $\phi_0(H_2, \underline{\lambda}^\star) > 0$, it follows that $P_0(H_1|\mathcal{H}_0) > 0$ and $P_0(H_2|\mathcal{H}_0) > 0$. Hence, conditions $(i)$ and $(ii)$ in (2) hold with $K = \mathcal{H}_0$. Hence, $\mathcal{X}_0$ is a $P_0-$quasi additive class. □

From Theorem 5 and Theorem 2, it immediately follows

**Corollary 1.** *The conditional probability $P_0$ on $\mathcal{E}_0 \times \mathcal{X}_0$ is coherent.*

We define the following partition of $\mathcal{F}_0$

$$\mathcal{F}^z_0 = \{E|H \in \mathcal{F}_0 : H \in D^z_0\}$$
$$\mathcal{F}^+_0 = \{E|H \in \mathcal{F}_0 : H \in D^+_0\}.$$

**Remark 4.** Note that as $\underline{\lambda}^\star$ is a solution of system $(\mathcal{S}_0)$ one has

$$P_0(E|H) = p^{(0)}_{E|H},\ \forall E|H \in \mathcal{F}^+_0\,. \tag{7}$$

If $\mathcal{F}_0^z \neq \emptyset$, setting $\mathcal{F}_1 = \mathcal{F}_0^z$, we consider the pair $(\mathcal{F}_1, \mathcal{A}_1)$. Note that, since $D_0^+$ cannot be empty it follows that $\mathcal{F}_1$ is a strict subset of $\mathcal{F}_0$. Then, repeating what has already been done to construct $P_0$, replacing index 0 by 1, we are able to define a real function $P_1$ on $\mathcal{E}_1 \times \mathcal{X}_1$ which is a coherent conditional probability on $\mathcal{E}_1 \times \mathcal{X}_1$, with $\mathcal{X}_1$ quasi additive w.r.t. $P_1$.
Consequently, if $\mathcal{F}_1^z \neq \emptyset$, we set $\mathcal{F}_2 = \mathcal{F}_1^z$ and so on. We continue in this way until we obtain $\mathcal{F}_k^z = \emptyset$ for some integer $k$. Therefore, we construct a strictly decreasing sequence $(\mathcal{F}_0, \mathcal{A}_0), (\mathcal{F}_1, \mathcal{A}_1), \ldots (\mathcal{F}_k, \mathcal{A}_k)$. Incidentally, we also have a sequence $\{\mathcal{F}_0^+, \mathcal{F}_1^+, \ldots, \mathcal{F}_k^+\}$ which is a partition of $\mathcal{F}_{J_n}$. In this way we are able to construct a finite sequence $P_0, P_1, \ldots, P_k$ such that, for each $i = 0, 1, \ldots, k$, $P_i$ is a conditional probability on $\mathcal{E}_i \times \mathcal{X}_i$, with $\mathcal{X}_i$ quasi additive w.r.t. $P_i$. Moreover, each $P_i$ is also coherent. Based on a reasoning as in Remark 4, for each $i = 0, 1, \ldots, k$, we have

$$P_i(E|H) = p_{E|H}^{(i)} \quad, \forall E|H \in \mathcal{F}_i^+ . \tag{8}$$

Finally, we note that the finite sequence $\{\mathcal{E}_0, \mathcal{E}_1, \ldots, \mathcal{E}_k\}$ is not increasing and given two sets $\mathcal{X}_i, \mathcal{X}_j$ in $\{\mathcal{X}_0, \mathcal{X}_1, \ldots, \mathcal{X}_k\}$, with $i \neq j$, one has

$$\mathcal{X}_i \cap \mathcal{X}_j = \emptyset. \tag{9}$$

**Example 1** (continued)
We set $(\mathcal{F}_0, \mathcal{A}_0) = (\mathcal{F}_{J_3}, \mathcal{A}_{J_3})$. The set $\mathcal{C}_0(\mathcal{H}_0)$ of the constituents associated with $\mathcal{F}_0$ and contained in $\mathcal{H}_0 = \{D \vee AC \vee AB\}$ is given in (1). By setting $\mathcal{P}_0 = (\frac{1}{2}, 0, \frac{1}{3})$ as a precise assessment on $\mathcal{F}_0$, it can be proved that $\mathcal{P}_0 \in \Pi_0$; moreover the system $(\mathcal{S}_0)$ associated with $(\mathcal{F}_0, \mathcal{P}_0)$ is

$$(\mathcal{S}_0) \begin{cases} \lambda_3 = \frac{1}{2}(\lambda_1 + \lambda_3 + \lambda_5 + \lambda_7) \\ \lambda_3 + \lambda_4 = 0(\lambda_3 + \lambda_4 + \lambda_5 + \lambda_6) \\ \lambda_3 + \lambda_4 = \frac{1}{3}(\lambda_1 + \lambda_2 + \lambda_3 + \lambda_4) \\ \lambda_1 + \lambda_2 + \lambda_3 + \lambda_4 + \lambda_5 + \lambda_6 + \lambda_7 = 1 \\ \lambda_h \geq 0, \ h = 1, 2, \ldots, 7 . \end{cases}$$

Coherence of $\mathcal{P}_0$ requires that system $(\mathcal{S}_0)$ is solvable. The vector $\underline{\lambda}^* = (\lambda_h^*, h = 1, \ldots, 7)$ with $\lambda_6^* = 1$ and $\lambda_h^* = 0$, if $h \neq 6$, is a solution of $(\mathcal{S}_0)$. Moreover, we have $\phi_0(D, \underline{\lambda}^*) = \lambda_1^* + \lambda_3^* + \lambda_5^* + \lambda_7^* = 0$, $\phi_0(AC, \underline{\lambda}^*) = \lambda_3^* + \lambda_4^* + \lambda_5^* + \lambda_6^* = 1$ and $\phi_0(AB, \underline{\lambda}^*) = \lambda_1^* + \lambda_2^* + \lambda_3^* + \lambda_4^* = 0$. Thus, $D_0^z = \{D, AB\}$ and $D_0^+ = \{AC\}$. Then, we set $P_0 : \mathcal{E}_0 \times \mathcal{X}_0$ as in (6), where $\mathcal{E}_0$ is the algebra generated by the constituents given in (1) and $\mathcal{X}_0 = \{AC, D \vee AC \vee AB\}$ is trivially a quasi additive class w.r.t. $P_0$. In particular, the value

$$P_0(E_2|H_2) = P_0(B|AC) = \frac{\lambda_3^* + \lambda_4^*}{\lambda_3^* + \lambda_4^* + \lambda_5^* + \lambda_6^*} = 0$$

is consistent with $\mathcal{A}_{J_3}$. Now, since $\mathcal{F}_0^+ = \{B|AC\}$, we set $\mathcal{F}_1 = \mathcal{F}_0 \setminus \{B|AC\}$ and we consider the pair $(\mathcal{F}_1, \mathcal{A}_1) = (\{ABC|D, C|AB\}, ([\frac{1}{2}, 1], [\frac{1}{3}, \frac{2}{3}]))$. The set $\mathcal{C}_1$ of the constituents associated with the family $\mathcal{F}_1$ is

$$\mathcal{C}_1 = \{R_0, R_2, R_4, R_5, R_6, R_7\} = \{C_1, \ldots, C_6\},$$

where $C_1 = ABC^cD$, $C_2 = ABC^cD^c$, $C_3 = ABCD$, $C_4 = ABCD^c$, $C_5 = (A^c \vee B^c)D, C_6 = (A^c \vee B^c)D^c$. Moreover $\mathcal{H}_1 = \{D \vee AB\}$ and $C_1(\mathcal{H}_1) = \mathcal{C}_1 \setminus \{C_6\}$. We choose the sub-assessment $(\frac{1}{2}, \frac{1}{3})$ of $\mathcal{P}_0$ on $\mathcal{F}_1$ as the precise coherent assessment $\mathcal{P}_1$ on $\mathcal{F}_1$ consistent with $\mathcal{A}_1$ (we could have chosen any other coherent assessment $\mathcal{P}_1$ on $\mathcal{F}_1$ consistent with $\mathcal{A}_1$). The system $(\mathcal{S}_1)$ associated with the pair $(\mathcal{F}_1, \mathcal{P}_1)$, where $\mathcal{P}_1 = (\frac{1}{2}, \frac{1}{3})$, is

$$(\mathcal{S}_1) \begin{cases} \lambda_3 = \frac{1}{2}(\lambda_1 + \lambda_3 + \lambda_5) \\ \lambda_3 + \lambda_4 = \frac{1}{3}(\lambda_1 + \lambda_2 + \lambda_3 + \lambda_4) \\ \lambda_1 + \lambda_2 + \lambda_3 + \lambda_4 + \lambda_5 = 1 \\ \lambda_h \geq 0, \ h = 1, 2, \ldots, 5 . \end{cases}$$

Since coherence of $\mathcal{P}_0$ requires coherence of each subvector, we have that the assessment $\mathcal{P}_1$ on $\mathcal{F}_1$ is coherent. Then, system $(\mathcal{S}_1)$ is solvable and a solution is given by $\underline{\lambda}^* = (\lambda_h^*, h = 1, \ldots, 5)$ with $\lambda_1^* = \lambda_2^* = \lambda_3^* = \frac{1}{3}$ and $\lambda_4^* = \lambda_5^* = 0$. We have $\phi_1(D, \underline{\lambda}^*) = \lambda_1^* + \lambda_3^* + \lambda_5^* = \frac{2}{3}$ and $\phi_0(AB, \underline{\lambda}^*) = \lambda_1^* + \lambda_2^* + \lambda_3^* + \lambda_4^* = 1$. Thus, $D_1^z = \emptyset$ and $D_1^+ = \{D, AB\}$. Then, we set $P_1 : \mathcal{E}_1 \times \mathcal{X}_1$ as in (6) where index 0 is replaced by index 1, $\mathcal{E}_1$ is the algebra generated by the constituents $C_1, \ldots, C_5$ and $\mathcal{X}_1$ is the class of conditional events $\{D, AB, D \vee AB\}$. In particular, the values

$$P_1(ABC|D) = \frac{\lambda_3^*}{\lambda_1^* + \lambda_3^* + \lambda_5^*} = \frac{\frac{1}{3}}{\frac{2}{3}} = \frac{1}{2},$$
$$P_1(C|AB) = \frac{\lambda_3^* + \lambda_4^*}{\lambda_1^* + \lambda_2^* + \lambda_3^* + \lambda_4^*} = \frac{\frac{1}{3}}{1} = \frac{1}{3}$$

are consistent with $\mathcal{A}_{J_3}$. Moreover, $\mathcal{X}_1$ is quasi additive w.r.t. $P_1$ ($\mathcal{X}_1$ is also additive); in particular $P_1(D|\mathcal{H}_1) + P_1(AB|\mathcal{H}_1) > 0$.

# 5 Extension of the probability $P_i$ on $\mathcal{E}_i \times \mathcal{X}_i$ to $P_i^0$ on $\mathcal{E}_0 \times \mathcal{X}_i$

In this section, for each $i = 1, \ldots, k$ (supposing $k \geq 1$), we introduce a conditional probability $P_i^0$ on $\mathcal{E}_0 \times \mathcal{X}_i$, with $\mathcal{X}_i$ quasi additive w.r.t. $P_i^0$, such that its restriction on $\mathcal{E}_i \times \mathcal{X}_i$ is the probability $P_i$ defined above. Rather than discuss this in full generality, let us look at $i = 1$. Based on definition (6), where we have replaced index 0 by 1, for every $E|H$ with $E \in \mathcal{E}_1$, $H \in \mathcal{X}_1$ the probability $P_1$ is defined as follows

$$P_1(E|H) = \frac{\phi_1(EH, \underline{\delta}^\star)}{\phi_1(H, \underline{\delta}^\star)}, \tag{10}$$

where $\underline{\delta}^\star$ is a solution of the following system $(\mathcal{S}_1)$ in the unknown $\underline{\delta} = (\delta_B, \ B \in \mathcal{C}_1)$

$$(\mathcal{S}_1) \begin{cases} \sum_{B \in \mathcal{C}_1(EH)} \delta_B = p_{E|H} \cdot \sum_{B \in \mathcal{C}_1(H)} \delta_B \,, \forall E|H \in \mathcal{F}_1 \\ \sum_{B \in \mathcal{C}_1} \delta_B = \sum_{B \in \mathcal{C}_1(\mathcal{H}_1)} \delta_B = 1 \,, \ \ \delta_B \geq 0 \ \ \forall B \in \mathcal{C}_1. \end{cases}$$

Since $\mathcal{E}_0 \supseteq \mathcal{E}_1$, each constituent $B \in \mathcal{C}_1$ can be written as the disjunction of the constituents contained in $\mathcal{C}_0(B)$, such as

$$B = \bigvee_{C \in \mathcal{C}_0(B)} C. \tag{11}$$

We note that, as $\mathcal{C}_0$ is a partition of $\Omega$, each set $\mathcal{C}_0(B)$ is non empty. Moreover, given two constituents $B'$ and $B''$ belonging to $\mathcal{C}_1$, as $B'B'' = \emptyset$, we have

$$\mathcal{C}_0(B') \cap \mathcal{C}_0(B'') = \emptyset. \tag{12}$$

Next, adding, for each component $\delta_B$ of $\underline{\delta}$, the following constraints

$$\delta_B = \sum_{C \in \mathcal{C}_0(B)} \lambda_C, \ \ \lambda_C \geq 0, \ \ \forall C \in \mathcal{C}_1(B) \tag{13}$$

to system $(\mathcal{S}_1)$, we obtain the following system $(S_1^0)$ in the unknown $\underline{\lambda} = (\lambda_C, \ \ C \in \mathcal{C}_0)$ associated with $(S_1)$

$$\begin{cases} \sum_{B \in \mathcal{C}_1(EH)} (\sum_{C \in \mathcal{C}_0(B)} \lambda_C) = p_{E|H} \cdot \sum_{B \in \mathcal{C}_1(H)} (\sum_{C \in \mathcal{C}_0(B)} \lambda_C), \\ \hfill \forall E|H \in \mathcal{F}_1, \\ \sum_{B \in \mathcal{C}_1} (\sum_{C \in \mathcal{C}_0(B)} \lambda_C) = \sum_{B \in \mathcal{C}_1(\mathcal{H}_1)} (\sum_{C \in \mathcal{C}_0(B)} \lambda_C) = 1, \\ \delta_B = \sum_{C \in \mathcal{C}_0(B)} \lambda_C, \ \forall B \in \mathcal{C}_1, \\ \lambda_C \geq 0, \quad \forall \ C \in \mathcal{C}_0(B) \,, \ \forall \ B \in \mathcal{C}_1. \end{cases}$$

Then, from (11) and (12) we deduce that for each solution of $(S_1)$ there exists at least a solution (in general infinite solutions ) of $(S_1^0)$. In particular, indicating by $r_0(\cdot)$ the cardinality of $\mathcal{C}_0(\cdot)$, a solution of $(S_1)$ could be $\underline{\lambda}^* = (\lambda_C^*, \ C \in \mathcal{C}_0)$ with

$$\lambda_C^* = \frac{\delta_C^\star}{r_0(B)} \,, \ \ \forall C \in \mathcal{C}_0(B) \,, \ \ B \in \mathcal{C}_1 \,. \tag{14}$$

Therefore, choosing a solution $\underline{\lambda}^*$ of $\mathcal{S}_1^0$ we set

$$P_1^0(E|H) = \frac{\phi_0(EH, \underline{\lambda}^*)}{\phi_0(H, \underline{\lambda}^*)}, \forall E|H, \ \text{with} \ E \in \mathcal{E}_0, \ H \in \mathcal{X}_1. \tag{15}$$

Then, we have

**Proposition 1.** *The restriction to $\mathcal{E}_1 \times \mathcal{X}_1$ of the function $P_1^0$ defined on $\mathcal{E}_0 \times \mathcal{X}_1$ is $P_1$.*

*Proof.* It is sufficient to prove that for every event $E \in \mathcal{E}_1$ it must be $\phi_1(E, \underline{\delta}^\star) = \phi_0(E, \underline{\lambda}^*)$. From (11) and (12) we have $C_0(E) = \bigvee_{B \in \mathcal{C}_1(E)} \left( \bigvee_{C \in \mathcal{C}_0(B)} C \right)$. Then, we obtain

$$\phi_1(E, \underline{\delta}^\star) = \sum_{B \in \mathcal{C}_1(E)} \delta_B^\star = \sum_{B \in \mathcal{C}_1(E)} \sum_{C \in \mathcal{C}_0(B)} \lambda_C^* = \sum_{C \in \mathcal{C}_0(E)} \lambda_C^* = \phi_0(E, \underline{\lambda}^*).$$

□

Finally, we have

**Theorem 6.** *The real function $P_1^0$ is a conditional probability on $\mathcal{E}_0 \times \mathcal{X}_1$. Moreover, the class $\mathcal{X}_1$ is quasi-additive w.r.t. $P_1^0$.*

*Proof.* By repeating the reasoning done in the first part of the proof of Theorem 5, it can be shown that $P_1^0$ is a conditional probability on $\mathcal{E}_0 \times \mathcal{X}_1$. In addition, we observe that quasi additive property involves only conditioning events in $\mathcal{X}_1 \subseteq \mathcal{E}_1$, where $\mathcal{X}_1$ is $P_1-$quasi additive. Then, from Proposition 1 it follows that $\mathcal{X}_1$ is a quasi additive class w.r.t. $P_1^0$. □

By repeating the reasoning above for $i = 2 \dots k$ and by setting $P_0^0 = P_0$ we are able to construct a sequence of conditional probabilities $P_i^0$ defined on $\mathcal{E}_0 \times \mathcal{X}_i$, $i = 0, \dots k$, such that, for each $i$, $\mathcal{X}_i$ is quasi additive w.r.t. $P_i^0$. Furthermore, by Theorem 2, each $P_i^0$ is a coherent conditional probability on $\mathcal{E}_0 \times \mathcal{X}_i$.

## 6 A quasi additive class with all conditioning events of $\mathcal{F}$.

In this section we will construct a conditional probability $P_{01\dots k}$ on $\mathcal{E} \times \mathcal{X}$, where $\mathcal{E} = \mathcal{E}_0$ and $\mathcal{X}$ is the union of the classes $\mathcal{X}_0, \mathcal{X}_1, \dots, \mathcal{X}_k$. Therefore, each conditioning event of $\mathcal{F}$ will belong to $\mathcal{X}$. Next, we will show that $\mathcal{X}$ is a quasi additive class respect to this probability. Finally, we will see that $P_{01\dots k}$ coincides with $\mathcal{P}$ on $\mathcal{F}$.
Let $\{P_i^0, \ i = 0, \dots, k\}$ be the sequence of conditional probabilities introduced in the previous section, with $P_i^0$ defined on $\mathcal{E}_0 \times \mathcal{X}_i$. For every $E|H$ with $E \in \mathcal{E}_0$ and $H \in \mathcal{X}$, where $\mathcal{X} = \mathcal{X}_0 \cup \mathcal{X}_1 \cup \dots \cup \mathcal{X}_k$, we set

$$P_{01\dots k}(E|H) = P_i^0(E|H), \ \ \text{with } i \text{ such that } H \in \mathcal{X}_i \tag{16}$$

**Remark 5.** The previous definition of $P_{01\dots k}$ is not ambiguous. In fact, given an event $H \in \mathcal{X}$, by relation (9) there exists a unique $\mathcal{X}_i$ in $\{\mathcal{X}_0, \mathcal{X}_1, \dots, \mathcal{X}_k\}$ such that $H \in \mathcal{X}_i$.

Then, we have

**Theorem 7.** *The real function $P_{01\ldots k}$ is a conditional probability on $\mathcal{E}_0 \times \mathcal{X}$. Moreover, the class $\mathcal{X}$ is quasi-additive w.r.t. $P_{01\ldots k}$.*

*Proof.* In order to prove that $P_{01\ldots k}$ is a conditional probability on $\mathcal{E}_0 \times \mathcal{X}$ we have to show that $P_{01\ldots k}$ satisfies properties $(i), (ii), (iii)$. Let $H$ be a conditioning event in $\mathcal{X}$, we have that $H \in \mathcal{X}_r$ for some $r = \{0, 1, \ldots, k\}$. Since both properties $(i)$ and $(ii)$ are known to hold for $P_r^0$, from definition (15) it follows that they also hold for $P_{01\ldots k}$. Next, given three events $E_1, E_2$ and $H$, with $E_1, E_2 \in \mathcal{E}_0$ and $H, E_1H \in \mathcal{X}$, the proof that $P_{01\ldots k}$ satisfies property $(iii)$ falls naturally into two cases.
*a*) If $H \in \mathcal{X}_r$ and $E_1H \in \mathcal{X}_r$ for some $r \in \{0, 1, \ldots, k\}$, then property $(iii)$ holds for $P_{01\ldots k}$ since it holds for $P_r^0$.
*b*) If $H \in \mathcal{X}_r$ and $E_1H \in \mathcal{X}_s$, for some $r, s \in \{0, 1, \ldots, k\}$ with $r < s$ , then it follows that $E_1H \notin \mathcal{X}_r$. Therefore, we have $P_r^0(E_1H|\mathcal{H}_r) = 0$. Moreover, as $E_1E_2H \subseteq E_1H$ one has $P_r^0(E_1E_2H|\mathcal{H}_r) = 0$. Thus, we obtain

$$P_{01\ldots k}(E_1E_2|H) = \frac{P_r^0(E_1E_2H|\mathcal{H}_r)}{P_r^0(H|\mathcal{H}_r)} = 0$$

and

$$P_{01\ldots k}(E_1|H) = \frac{P_r^0(E_1H|\mathcal{H}_r)}{P_r^0(H|\mathcal{H}_r)} = 0.$$

Then, property $(iii)$, that is $P_{01\ldots k}(E_1E_2|H) = P_{01\ldots k}(E_2|E_1H)P_{01\ldots k}(E_1|H)$, is satisfied by $0 = 0$.
To prove that $\mathcal{X}$ is a quasi additive class w.r.t.$P_{01\ldots k}$ we have to show that, the conditions $(i)$ and $(ii)$ in (2) are satisfied by $P_{01\ldots k}$. Let $H_1$ and $H_2$ be two conditioning events in $\mathcal{X}$, we distinguish two cases:
*a*) $H_1, H_2 \in \mathcal{X}_r$, for some $r \in \{0, 1, \ldots, k\}$. Quasi additivity conditions $(i)$ and $(ii)$ in (2) hold for $P_{01\ldots k}$ since they hold for $P_r^0$;
*b*) $H_1 \in \mathcal{X}_r$ and $H_2 \in \mathcal{X}_s$ for some $r, s \in \{0, 1, \ldots, k\}$. In this case, we consider $\mathcal{H}_m$, where $m = \min\{r, s\}$. Since $\mathcal{H}_m$ is the disjunction of the conditioning events contained in $D_m$, it follows that $\mathcal{H}_m \in \mathcal{X}_m$. Moreover, we have

$$H_1 \vee H_2 \subseteq \bigvee_{H \in D_m} H = \mathcal{H}_m,$$

then condition $(i)$ in (2) is satisfied by $K = \mathcal{H}_m$. Next, we have

$$P_{01\ldots k}(H_1|\mathcal{H}_m) + P_{01\ldots k}(H_2|\mathcal{H}_m) > 0.$$

In fact, if $m = r$, then $P_{01\ldots k}(H_1|\mathcal{H}_m) = P_r^0(H_1|\mathcal{H}_r) > 0$, otherwise if $m = s$, then $P_{01\ldots k}(H_2|\mathcal{H}_m) = P_s^0(H_2|\mathcal{H}_s) > 0$. We can conclude that condition $(ii)$ in (2) is satisfied by $K = \mathcal{H}_m$. □

**Remark 6.** We observe that in the first part of the previous proof the further case $H \in \mathcal{X}_r$ and $E_1H \in \mathcal{X}_s$, with $r > s$, cannot be possible. In fact, if it were true, as $H \notin \mathcal{X}_s$, there would be $P_s(H|\mathcal{H}_s) = 0$. Therefore, as $E_1H \subseteq H$ we would have

$$0 = P_s^0(H|\mathcal{H}_s) \geq P_s^0(E_1H|\mathcal{H}_s) > 0,$$

which is absurd.

Finally, based on Theorem 7 and Theorem 2 we can conclude that $P_{01\ldots k}$ is a coherent probability on $\mathcal{E}_0 \times \mathcal{X}$. Moreover, given a conditional event $E|H \in \mathcal{F}_{J_n}$ as $(\mathcal{F}_0^+, \mathcal{F}_1^+, \ldots, F_k^+)$ is a partition of $\mathcal{F}_{J_n}$ there exist $i \in \{0, 1, \ldots, k\}$ such that $E|H \in \mathcal{F}_i^+$. Then, from (15) and (8), for each $i = 0, 1, \ldots, k$ it follows that

$$P_{01\ldots k}(E|H) = p_{E|H}^{(i)}, \ \forall E|H \in \mathcal{F}_i^+ \, .$$

Therefore $P_{01\ldots k}$ is consistent with $\mathcal{A}_{J_n}$.

**Remark 7.** Since any restriction of a coherent conditional probability is coherent too, then the restriction of $P_{01\ldots k}$ on $\mathcal{F}_{J_n}$ is coherent. We observe that $P_{01\ldots k}$ is, in general, not unique, although we start from the same precise coherent assessment. Moreover, two obtained conditional probabilities $P_{01\ldots k}$ and $P'_{01\ldots k}$ could be defined on different sets of conditioning events.

**Example 1** (continued)
We set $P_0^0 = P_0$ and, by applying (15), we extend the conditional probabilities $P_1 : \mathcal{E}_1 \times \mathcal{X}_1$ to $P_1^0$ on $\mathcal{E}_0 \times \mathcal{X}_1$. Then, the function $P_{01} : \mathcal{E}_0 \times \mathcal{X}$ defined as

$$P_{01}(E|H) = \begin{cases} P_0^0(E|H), \text{ if } H \in \{AC, D \vee AC \vee AB\} \\ P_1^0(E|H), \text{ if } H \in \{D, AB, D \vee AB\}, \end{cases}$$

where $\mathcal{X} = \{AC, D \vee AC \vee AB, D, AB, D \vee AB\}$, is a coherent conditional probability on $\mathcal{E}_0 \times \mathcal{X} \supseteq \mathcal{F}$ Moreover, we have that: (a) $P_{01}$ is consistent with $\mathcal{A}_{J_3}$; (b) $P_{01}(E|H) = p_{E|H}$ for every $E|H \in \mathcal{F}_{J_3}$, that is

$$P_{01}(ABC|D) = \tfrac{1}{2}, \ P_{01}(B|AC) = 0, P_{01}(C|AB) = \tfrac{1}{3} \ ;$$

(c) $\mathcal{X}$ is $P_{01}$-quasi additive. We observe that the class $\mathcal{X}$ is not additive (for instance $D \vee AC \notin \mathcal{X}$).

## 7 Further results

In this section we will give a characterization theorem of coherent precise assessment on a finite family of conditional events and a relevant way of introducing quasi additive classes.
Let $\mathcal{P}$ be a precise coherent assessment on a finite family $\mathcal{F}_{J_n}$ of conditional events. We recall that $\mathcal{P}$ is a particular interval-valued probability assessment on $\mathcal{F}_{J_n}$ where each upper bound coincides with the respective lower bound, then by setting $\mathcal{A}_{J_n} = \mathcal{P}$ and based on the results obtained in previous we are able to construct (in a direct way) a coherent conditional probability $P_{01\dots k}$ on $\mathcal{E}_0 \times \mathcal{X} \supseteq \mathcal{F}_{J_n}$, with $\mathcal{X}$ quasi additive w.r.t. $P_{01\dots k}$, such that the restriction of $P_{01\dots k}$ on $\mathcal{F}_{J_n}$ coincides with $\mathcal{P}$. Moreover, as any restriction of a coherent conditional probability is coherent too, we have

**Theorem 8.** *A real-valued function $\mathcal{P}$ on a finite family of conditional events $\mathcal{F}_{J_n}$ is coherent if and only if $\mathcal{P}$ can be extended as a conditional probability $P_{01\dots k}$ on $\mathcal{E}_0 \times \mathcal{X} \supseteq \mathcal{F}_{J_n}$, with $\mathcal{X}$ quasi additive w.r.t. $P_{01\dots k}$.*

A similar result, which also has been proved when the family of conditional events is infinite, has been given in Coletti and Scozzafava [2002] (see also Coletti and Scozzafava [1999]) by assuming $\mathcal{X}$ additive. Moreover, given a precise coherent assessment $\mathcal{P}$ on a finite family of conditional events $\mathcal{F}_{J_n}$, let $\mathcal{D}$ be the associated set of conditioning events. By the results of the previous section we can show that the cardinality of the quasi additive set $\mathcal{X} \supseteq \mathcal{D}$ w.r.t. any coherent extension $P_{01\dots k}$ is always at most $2n$, where $n$ denotes the cardinality of $\mathcal{D}$. We observe that for an additive set $\mathcal{X}' \supseteq \mathcal{D}$ the cardinality could be at most $2^n - 1$.

**Remark 8.** The problem of giving an algorithm that characterizes the whole set of conditional probabilities consistent with the initial imprecise assessment seems not easy and we do not consider it in this paper. However, we illustrate a procedure for constructing relevant cases of conditional probabilities by considering the algorithm for checking g-coherence of imprecise assessments given in Biazzo and Gilio [2000]. Based on (3), denoting by $\Lambda_0$ the set of solutions of system $S_0$, we introduce the following sets of conditional events

$$I_0 = \{H \in D_0 : \phi_0(H, \underline{\lambda}) = 0,\ \forall \underline{\lambda} \in \Lambda_0\}\,,$$
$$\Gamma_0 = D_0 \setminus I_0\,.$$

We observe that $I_0$ is equivalent to the set, denoted by the same symbol, used in the algorithm for checking g-coherence given in Biazzo and Gilio [2000]. As for any given $H \in I_0$ there exists a solution $\underline{\lambda}_H$ such that $\phi_0(H, \underline{\lambda}_H) > 0$, then (by a convex linear combination of the vectors $\underline{\lambda}_H$'s, with coefficients all positive) there exists a solution $\underline{\lambda}$ such that $\phi_0(H, \underline{\lambda}) > 0$ for all $H \in I_0$. Therefore, by recalling (5), $\Gamma_0$ is the set $D_0^+$ associated with the solution $\underline{\lambda}$, while in general the sets $D_0^+, D_0^z$ associated with any other solution $\underline{\lambda}^* \neq \underline{\lambda}$ are such that $\Gamma_0 \supseteq D_0^+$ and $I_0 \subseteq D_0^z$. Then, for each $I \subseteq \Gamma_0$ and for each $J$ such that $I \subseteq J \subseteq D_0$, the class $I \cup \{\mathcal{H}_J\}$, where $\mathcal{H}_J = \bigvee_{H \in J} H$, is a quasi additive class which can be introduced at the first step of the procedure. Similar quasi additive classes can be introduced in the other steps of the procedure; at the end, based on the union of these quasi additive classes, we obtain a conditional probability consistent with the initial imprecise assessment.

We can apply the procedure above to any subset $\Lambda_0'$ of the set of solutions $\Lambda_0$ of the system $S_0$; in this way we can determine other quasi additive classes of conditioning events and the associated conditional probabilities consistent with the initial assessment.
Concerning the computational complexity, we observe that the procedure to construct the coherent probability $P_{01\dots k}$ is related to the problem of the global checking of coherence of the initial assessment, which tends to become intractable when the cardinality of the starting family of conditional events increases (for an analysis of complexity on this kind of problems see e.g. Biazzo et al. [2005]). Local methods for reducing the computational difficulties have been developed in some papers (see e.g. Biazzo et al. [2003], Capotorti and Vantaggi [2002], Capotorti et al. [2003]).

**Example 2.** Let be given an algebra of event $\mathcal{E}_0$ and a probability $P_0$ on $\mathcal{E}_0$. Of course, $P_0$ is coherent. Moreover, let $\mathcal{X} = \{H_1, H_2, \dots, H_n, \Omega\}$ be any subset of $\mathcal{E}_0 \setminus \{\emptyset\}$, with $P_0(H_i) > 0$, $i = 1, \dots, n$. The probability $P_0$ can be extended to a conditional probability $P$ on $\mathcal{E}_0 \times \mathcal{X}$ defined as

$$P(E|H_i) = \frac{P_0(EH_i)}{P_0(H_i)},\ \forall E \in \mathcal{E}, H_i \in \mathcal{X},$$

with $P(E|\Omega) = P_0(E)$, for every $\mathcal{E}$. We remark that: $(i)$ $H_i \vee H_j \subseteq \Omega$ for every subset $\{i, j\}$ of $\{1, 2, \dots, n\}$; $(ii)$ $P(H_i|\Omega) + P(H_j|\Omega) > 0$; thus, $\mathcal{X}$ is $P$-quasi additive and, of course, the conditional probability $P$ is coherent. As we can see, the property of quasi additivity is implicitly exploited in all the cases in which we construct a (coherent) conditional probability by means of ratios of unconditional probabilities.

# 8 Conclusions

In this paper starting from a g-coherent interval-valued probability assessment $\mathcal{A}_{J_n}$ on a finite family $\mathcal{F}_{J_n}$ we first have constructed a sequence of conditional probabilities with quasi additive classes of conditioning events. Then we have extended each probability in this sequence to obtain a new sequence of conditional probabilities with quasi additive classes of conditioning events. Moreover, we have defined a conditional probability $P_{01\dots k}$ on $\mathcal{E} \times \mathcal{X}$, where $\mathcal{E}$ is the algebra generated by the constituents associated with $\mathcal{F}_{J_n}$ and $\mathcal{X}$ is a quasi additive class of conditioning events which contains all conditioning events of $\mathcal{F}_{J_n}$. We have shown that $P_{01\dots k}$ is coherent and consistent with the imprecise assessment $\mathcal{A}_{J_n}$ on $\mathcal{F}_{J_n}$. Finally, we have given a characterization theorem of coherent precise assessments and a relevant way of introducing quasi additive classes.

**Acknowledgements**

Thanks to the reviewers for helpful comments and suggestions.